\documentclass[10pt]{article}

\usepackage[utf8]{inputenc}
\usepackage[T1]{fontenc}
\usepackage[margin=1in]{geometry}
\usepackage{newtxtext,newtxmath}
\usepackage{graphicx}
\usepackage{booktabs}
\usepackage{amsmath}
\usepackage{array}
\usepackage[hidelinks]{hyperref}
\usepackage{caption}
\usepackage{enumitem}
\usepackage{microtype}
\usepackage{titlesec}
\usepackage{indentfirst}
\usepackage{authblk}

\setlist[itemize]{leftmargin=1.5em,itemsep=1pt,topsep=2pt}
\titlespacing{\section}{0pt}{7pt}{3pt}
\titlespacing{\subsection}{0pt}{5pt}{2pt}
\title{Density-Aligned Synthetic Training and Class-aware Ensemble\\
for Sim2Real Roadside LiDAR 3D Detection}

\newcommand{\teamname}{\textbf{Team Name:} luopuu}

\author[1]{Pu Luo}
\author[1]{Cong Xu}
\author[1]{Yumei Li}
\author[1]{Kexin Zhang}
\author[1]{Licheng Jiao}
\author[1]{Wenping Ma}
\author[1]{Lingling Li}

\affil[1]{Xidian University, Xi'an, China\newline
\footnotesize
Pu Luo: \texttt{25171214094@stu.xidian.edu.cn} \quad
Cong Xu: \texttt{25171214052@stu.xidian.edu.cn}\newline
Yumei Li: \texttt{25171213938@stu.xidian.edu.cn} \quad
Kexin Zhang: \texttt{kxzh@stu.xidian.edu.cn}\newline
Licheng Jiao: \texttt{lchjiao@mail.xidian.edu.cn} \quad
Wenping Ma: \texttt{wpma@mail.xidian.edu.cn}\newline
Lingling Li: \texttt{llli@xidian.edu.cn}\newline
\small \teamname
}
\date{}

\begin{document}
\maketitle
\vspace{-1.1em}

\begin{abstract}
We present our solution to the LUMPI track of the UCF UrbanTwin Sim2Real LiDAR Challenge at the 6th DriveX Workshop, ECCV 2026. The detector must be trained only on synthetic data and is evaluated on 50 held-out real LiDAR frames; a separate 50-frame synthetic submission is evaluated for point-cloud realism. Our method addresses the Sim2Real gap at three levels. First, we align synthetic scans to the 50k-point test density and build a 30k-record training pool using UT-LUMPI geometry, RangeLDM-based sampling diversification, rare-class copy-paste, and pedestrian-oriented augmentation. Second, complementary DSVT detectors and Car/Bus PointPillars specialists are trained under the same synthetic-only constraint. Third, predictions are integrated by class-aware routing, asymmetric agreement fusion, constrained residual-recall supplementation, class-coverage auditing, and selective box-size calibration. The realism branch is optimized independently with radial-density matching, weak affine calibration, and calibrated set mixing. The final submission obtains a Combined Score of 0.4692, a Detection Score of 0.1797, a Realism Score of 0.9035, and 3D mAP@0.5 of 0.1258.
\end{abstract}

\noindent\textbf{Keywords:} Sim2Real; roadside LiDAR; 3D object detection; synthetic data; density alignment; class-aware ensemble

\section{Introduction}
Roadside LiDAR provides stable 3D observations for infrastructure-side perception, but collecting and annotating real data at scale is costly. LUMPI is a multi-perspective roadside sensing dataset recorded at a large intersection in Hannover, Germany~\cite{lumpi}. UrbanTwin provides a high-fidelity synthetic counterpart, UT-LUMPI, generated from a digital twin of the same target environment~\cite{urbantwin}. These resources make LUMPI a suitable testbed for studying synthetic-to-real transfer.

The challenge nevertheless remains difficult because geometric similarity alone does not remove the observation gap. Synthetic and real scans differ in point count, radial density, local sampling, object morphology, and class frequency. These discrepancies change occupied-voxel statistics and per-object point support before learned features are formed. In addition, the eight evaluation classes---Person, Car, Bicycle, Motorcycle, Bus, Truck, Van, and Unknown---have very different frequencies and error modes.

Our design therefore follows a simple principle: \emph{first align the input distribution, then create controlled model diversity, and finally fuse only complementary evidence}. The detector is trained only with synthetic data. Public real data outside the challenge's forbidden-frame list are used only for permitted validation, model selection, and post-processing analysis; no forbidden frame is used for training, validation, tuning, or calibration. This follows the challenge protocol~\cite{challenge}.

Our main contributions are: (1) 50k point-density alignment and multi-source synthetic training; (2) role-based detector diversity using DSVT~\cite{dsvt} and class-specialized PointPillars~\cite{pointpillars}; (3) a class-aware fusion strategy designed around correlated errors and residual recall; and (4) a separately optimized realism branch with metric-level bottleneck analysis.

\section{Task and System Overview}
The LUMPI track evaluates 3D detection and synthetic-point-cloud realism. Detection uses class-averaged 3D AP at IoU 0.5 with KITTI 40-point interpolation as the primary metric; AP@0.7 is also reported. The 50 submitted synthetic frames are compared with 50 held-out real reference frames after masking to $[-40,40]\times[-40,40]\times[-5,5]$ m and uniformly sampling 10,000 points per side. Realism is computed from normalized Chamfer Distance (CD), Maximum Mean Discrepancy (MMD), Earth Mover's Distance (EMD), and Fr\'echet Point-cloud Distance (FPD). The final score is
\begin{equation}
S_{\mathrm{combined}}=0.6S_{\mathrm{det}}+0.4S_{\mathrm{real}}.
\end{equation}

\begin{figure*}[t]
    \centering
    \includegraphics[width=0.98\textwidth]{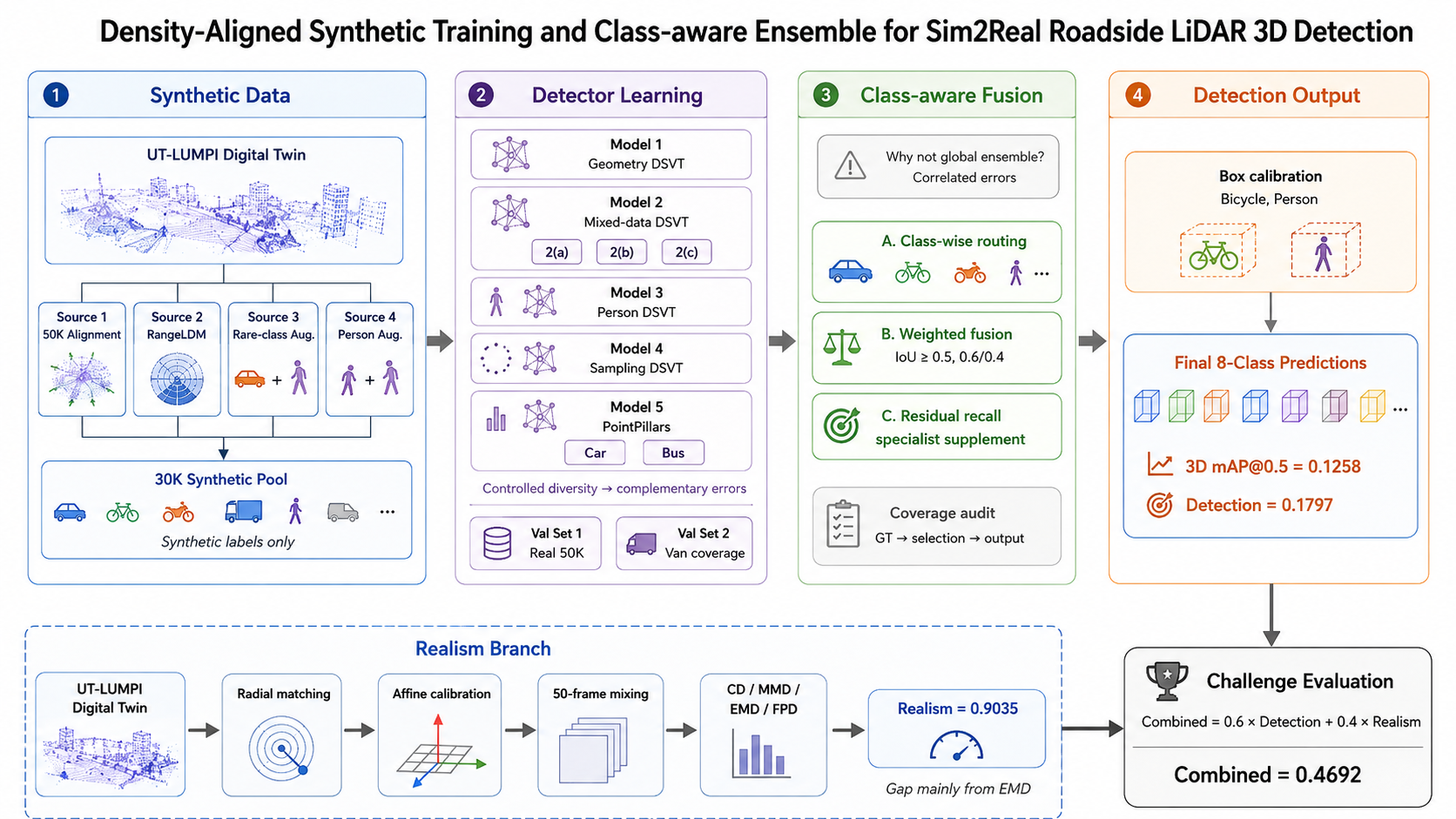}
    \caption{Overall framework. UT-LUMPI provides the geometric foundation. Four synthetic data sources support complementary detector learning; class-aware fusion addresses correlated errors and class-dependent domain gaps. The realism branch is optimized independently.}
    \label{fig:framework}
\end{figure*}

Figure~\ref{fig:framework} summarizes the method. The upper branch constructs density-aligned synthetic data, trains complementary detectors, performs class-aware fusion, and calibrates selected box geometries. The lower branch independently optimizes the 50 realism frames. Keeping these branches separate is important: detection benefits from category and observation diversity, whereas realism rewards global distributional similarity.

\section{Density-Aligned Multi-source Synthetic Training}
\subsection{Four Data Sources}
UT-LUMPI contains about 10k annotated frames and roughly 220k points per frame, while each released detection-test frame contains about 50k points. We therefore treat the digital twin as the geometric foundation and construct four detection-oriented sources consistent with Fig.~\ref{fig:framework}.

\textbf{Source 1: 50k density alignment.} Synthetic scans are randomly resampled to exactly 50,000 points. This is not merely an efficiency choice: with the same ensemble, our 150-frame real validation set scores 0.1464 mAP at its original $\sim$215k density but 0.1320 after 50k resampling. The former therefore overestimates performance under the released test condition. All subsequent selection and post-processing use the 50k protocol.

\textbf{Source 2: RangeLDM sampling diversification.} RangeLDM~\cite{rangeldm} generates LiDAR in range view using latent diffusion. We use RangeLDM-based synthetic scans as an additional sampling domain rather than replacing UT-LUMPI geometry. This exposes the detector to sparser and less regular return patterns while retaining a geometry-stable digital-twin source.

\textbf{Source 3: rare-class augmentation.} Bicycle, Motorcycle, Bus, Truck, and Van instances are extracted from synthetic labels and pasted into valid positions after random pose perturbation. This increases supervision for classes that would otherwise contribute few training instances despite having equal weight in class-averaged mAP.

\textbf{Source 4: pedestrian-oriented augmentation.} Person instances receive dedicated scale, orientation, and position perturbations to increase the diversity of small, sparsely sampled pedestrian observations.

The final training pool contains approximately 30k training records: an 8k density-aligned base stream, 6k RangeLDM-based scans, 2k rare-class copy-paste scans, a 10k density-aligned UT-LUMPI stream, and 4k pedestrian-oriented scans. Synthetic GT database sampling further balances class frequency. Increasing rare-class sampling beyond the final setting produced no additional gain, indicating that geometric and observation diversity, rather than raw repetition, became the dominant limitation.

\section{Complementary Detectors and Class-aware Fusion}
\subsection{Role-based Detector Learning}
Our main detector family uses DSVT~\cite{dsvt} with a unified 50k preprocessing pipeline. After harmonizing synthetic annotations to the challenge taxonomy, all models predict the eight evaluation categories but emphasize different synthetic distributions. \textbf{Model 1} is a geometry-stable DSVT trained mainly on density-aligned scans. \textbf{Model 2} is the mixed-data DSVT trained on the full multi-source pool; three stable late-stage members, denoted 2(a)--2(c), provide low-cost training-trajectory diversity. \textbf{Model 3} emphasizes pedestrian and sparse small-object observations. \textbf{Model 4} emphasizes sampling-diversified scans. \textbf{Model 5} contains Car and Bus PointPillars specialists~\cite{pointpillars}, providing an architecture different from the DSVT family. All submitted detector weights satisfy the synthetic-only training requirement.

\subsection{Why Class-aware Fusion}
A uniform ensemble is ineffective for two reasons. First, reliability is class dependent: small vulnerable road users, frequent cars, and low-frequency large vehicles exhibit different failure modes. Second, DSVT models share substantial representation and therefore make correlated errors. Adding every model to every class mainly increases duplicate predictions and false positives.

We instead assign a primary and auxiliary source per class using the density-aligned validation set. Candidate ordered pairs are evaluated by class, so a model is selected for complementarity rather than standalone mAP. For matched same-class boxes with
\begin{equation}
\mathrm{IoU}_{3D}(b_i,b_j)\geq0.5,
\end{equation}
we keep the primary box geometry and update only confidence:
\begin{equation}
s_{\mathrm{fuse}}=0.6s_{\mathrm{primary}}+0.4s_{\mathrm{aux}}.
\end{equation}
Unmatched candidates are retained to avoid unnecessary recall loss. Because the fusion is asymmetric, the primary/auxiliary order is also validated.

\subsection{Residual Recall, Coverage Audit, and Geometry Calibration}
Two-model routing resolves most class-dependent disagreement, but some categories still contain repeatable residual misses. Additional specialists are therefore admitted only when they contribute independent correct detections beyond the primary fusion. Their boxes are subject to IoU-based duplicate suppression and confidence discounting. In practice this mechanism is most useful for Car and Bus, where a specialist can recover misses without replacing the main DSVT geometry.

Class-wise selection also requires explicit coverage auditing. Our 150-frame real validation subset contains no Van ground truth. An early automatic selector therefore had no Van AP and omitted the class from the fusion map, even though base models produced Van predictions. We correct this with the logical check
\begin{equation}
\text{GT coverage}\rightarrow\text{selection coverage}\rightarrow\text{output coverage},
\end{equation}
and use a 1,000-frame synthetic validation set containing 4,282 Van boxes for relative Van model selection.

Finally, box geometry is calibrated only for classes with a stable synthetic-to-real size bias. Bicycle and Person benefit from class-specific dimension calibration estimated from permitted public training data outside the forbidden set. The same operation is disabled for Van because it sharply degrades validation performance. Thus, calibration is treated as a validated class-specific correction, not a universal post-processing rule.

\section{Independent Realism Optimization}
The 50 realism frames are optimized separately from the detector-training pool. First, points are reweighted by radial distance $r=\sqrt{x^2+y^2}$ so that the synthetic near/far density profile better matches permitted public LUMPI training data. Second, weak axis-wise scale and translation candidates are evaluated to correct low-order statistics without distorting the digital-twin layout. Finally, two stable calibrated variants are mixed at approximately 75\%:25\% to form the 50-frame submission.

The final Realism Score is 0.9035, with raw metrics CD=2.1805, MMD=0.000725, EMD=1.857, and FPD=0.3395. Our internal metric decomposition shows that most of the remaining penalty comes from EMD. As a diagnostic, independent 10k-point subsamples from permitted public real data produced EMD values on the order of 0.9--1.0, indicating substantial sampling variability at this scale. Additional global rotation, translation, and anisotropic-scaling searches did not materially reduce synthetic-to-real EMD. We therefore stopped low-return global calibration and prioritized detection.

\section{Experiments}
\subsection{Validation Protocol and Ablation}
\textbf{Validation Set 1} contains 150 labeled public real frames outside the forbidden list and is downsampled to 50k points for model selection and post-processing. \textbf{Validation Set 2} contains 1,000 synthetic frames and supplies Van coverage. Real labels are used only for permitted validation and strategy selection; they do not update detector weights.

Table~\ref{tab:ablation} reports the main detection improvements under the target-consistent 50k protocol. The 0.1464 score measured on the original high-density validation scans is only a density diagnostic and is not used as the method baseline.

\begin{table}[t]
\centering
\small
\begin{tabular}{p{6.7cm}c}
\toprule
\textbf{Configuration} & \textbf{mAP}\\
\midrule
50k density-aligned validation baseline & 0.1320\\
+ class-wise primary/auxiliary routing & 0.1413\\
+ Bicycle/Person geometry calibration & 0.1548\\
+ ordered-pair complementarity search & 0.1560\\
+ constrained residual recall and coverage audit & \textbf{0.1584}\\
\bottomrule
\end{tabular}
\caption{Detection ablation on the 50k validation protocol.}
\label{tab:ablation}
\end{table}

The improvement is therefore not explained by repeatedly increasing ensemble size. Each stage addresses a distinct problem: target-density mismatch, class-dependent model reliability, systematic box geometry bias, correlated model errors, and uncovered residual misses/classes.

\subsection{Final Result}
The hidden-test result is summarized in Table~\ref{tab:final}. The Detection Score is consistent with the reported raw mAP after benchmark normalization, and the Combined Score is consistent with the challenge weighting in Eq.~(1).

\begin{table}[t]
\centering
\small
\begin{tabular}{lc@{\qquad}lc}
\toprule
\textbf{Metric} & \textbf{Value} & \textbf{Metric} & \textbf{Value}\\
\midrule
Combined Score & \textbf{0.4692} & Detection Score & \textbf{0.1797}\\
3D mAP@0.5 & \textbf{0.125786} & 3D mAP@0.7 & 0.0559\\
BEV mAP & 0.1315 & Realism Score & \textbf{0.9035}\\
\bottomrule
\end{tabular}
\caption{Final hidden-test performance.}
\label{tab:final}
\end{table}

\section{Conclusion}
We developed a Sim2Real roadside LiDAR system centered on target-density alignment and class-dependent complementarity. A 30k-record synthetic pool combines UT-LUMPI geometry with sampling diversification and long-tail augmentation; role-based DSVT models and PointPillars specialists provide controlled prediction diversity; and class-aware routing, agreement fusion, residual-recall supplementation, coverage auditing, and selective geometry calibration address distinct error sources. A separate realism branch optimizes global point-cloud statistics without constraining detector training. The final system achieves a Combined Score of 0.4692 with Detection Score 0.1797 and Realism Score 0.9035. The main practical lesson is that low-level observation alignment and error-aware model complementarity can be more valuable than simply enlarging either the training set or the ensemble.

\end{document}